%% file: main.tex
\documentclass[10pt,twocolumn,letterpaper]{article}

\usepackage{iccv}
\usepackage{times}
\usepackage{epsfig}
\usepackage{graphicx}
\usepackage{amsmath}
\usepackage{amssymb}
\usepackage{amsfonts}
\usepackage{bm}
\usepackage{booktabs}
\usepackage{xcolor}
\usepackage{float}
\usepackage{subcaption}
\usepackage{balance}
\usepackage{enumitem}

\usepackage[pagebackref=true,breaklinks=true,letterpaper=true,colorlinks,bookmarks=false]{hyperref}

\iccvfinalcopy 

\def\iccvPaperID{6391} 

\newcommand{\rpm}{\raisebox{.3ex}{$\scriptstyle\pm$}}

\newcommand{\edit}[1]{{#1}}

\begin{document}

\title{DiffDreamer: Towards Consistent Unsupervised Single-view Scene Extrapolation with Conditional Diffusion Models}

\author{
Shengqu Cai$^{1,2*}$
\qquad Eric Ryan Chan$^{1}$
\qquad Songyou Peng$^{2,3}$
\qquad Mohamad Shahbazi$^{2}$
\\
\qquad Anton Obukhov$^{2}$ 
\qquad Luc Van Gool$^{2,4}$
\qquad Gordon Wetzstein$^{1}$
\\
$^{1}$Stanford University\quad
$^{2}$ETH Z\"urich\quad
$^{3}$MPI for Intelligent Systems, T\"ubingen \quad
$^{4}$KU Leuven
}


\input{fig/nerfdreamer/teaser.tex}
\begin{abstract}
Scene extrapolation---the idea of generating novel views by flying into a given image---is a promising, yet challenging task. \edit{
For each predicted frame, a joint inpainting and 3D refinement problem has to be solved, which is ill posed and includes a high level of ambiguity. 
Moreover, training data for long-range scenes is difficult to obtain and usually lacks sufficient views to infer accurate camera poses.} We introduce \emph{DiffDreamer}, an unsupervised framework capable of synthesizing novel views depicting a long camera trajectory while training solely on internet-collected images of nature scenes. \edit{Utilizing the stochastic nature of the guided denoising steps, we train the diffusion models to refine projected RGBD images but condition the denoising steps on multiple past and future frames for inference.} We demonstrate that image-conditioned diffusion models can effectively perform long-range scene extrapolation while preserving consistency significantly better than prior GAN-based methods.
\edit{DiffDreamer is a powerful and efficient solution for scene extrapolation, producing impressive results despite limited supervision.}
Project page: \url{https://primecai.github.io/diffdreamer}.
\end{abstract}

\input{sec/1_introduction}
\input{sec/2_related}
\input{sec/3_method}

\input{sec/4_results}
\input{sec/5_conclusions}

{
\small
\balance
\bibliographystyle{ieee_fullname}
\bibliography{main}
}

\input{sec/X_supplementary}

\end{document}

%% file: fig/nerfdreamer/teaser.tex
\twocolumn[{%
  \renewcommand\twocolumn[1][]{#1}%
\maketitle
\begin{center}
  \newcommand{\teaserwidth}{1.0\textwidth}
  \centerline{
    \includegraphics[width=\teaserwidth]{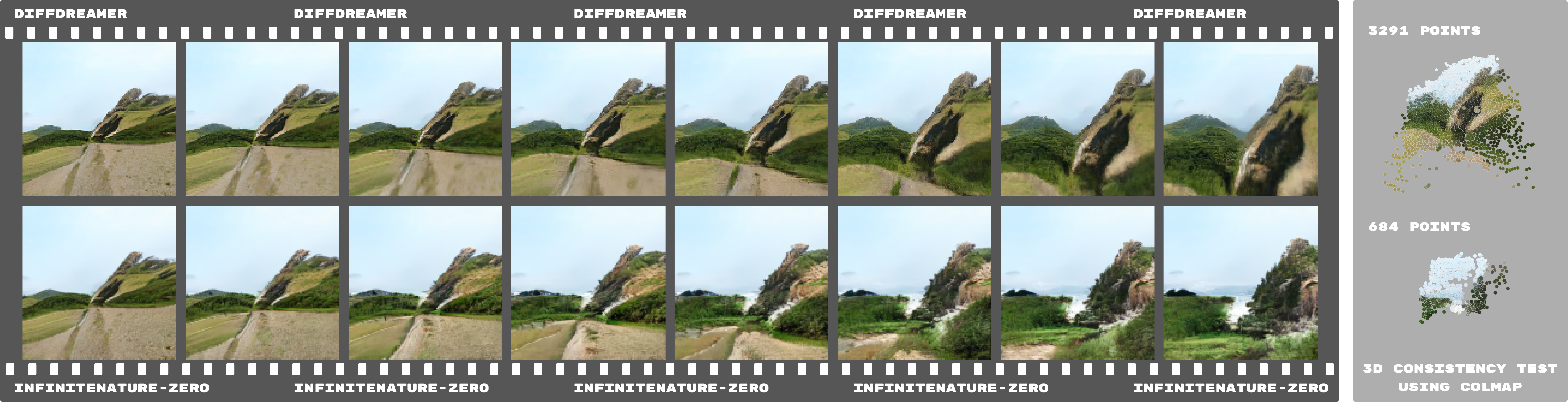}
  }
  \captionsetup{type=figure}
  \captionof{figure}{
    \textbf{DiffDreamer} (Top) is a novel diffusion-based approach for scene extrapolation. It exhibits high spatio-temporal consistency, a desired property missing in prior art, such as InfNat-0~\cite{li2022infnat_zero} (Bottom). 
    We check for consistency by extracting keypoints from the sequences with COLMAP, resulting in point clouds of vastly different sizes and sparsity (Right).
  }
  \vspace{-0.025in}
  \label{fig:teaser}
 \end{center}%
}]

%% file: sec/1_introduction.tex
\section{Introduction}
\label{sec:intro}
\let\thefootnote\relax\footnote{Corresponding author: Shengqu Cai (\href{mailto://shengqu@stanford.edu}{shengqu@stanford.edu}).}
3D content creation tools are the foundation of emerging metaverse applications, among many others. 
Current approaches primarily rely on heavy manual labor, making the process expensive and inefficient. 
We set out to make 3D content creation automated and accessible.
More specifically, an important downstream task we approach is \textit{consistent scene extrapolation}.
Given a single image and a long camera trajectory flying into the scene, the goal of \textit{consistent scene extrapolation} is to synthesize a multiview-consistent 3D scene along the camera trajectory. 
In other words, we want to teach a machine to hallucinate content when flying into the image while maintaining multiview consistency, thereby extrapolating the scene realistically.
Successfully addressing this task opens up a wide range of potential applications in virtual reality, 3D content creation, synthetic data creation, and 3D viewing platforms. 

Consistent scene extrapolation is extremely challenging as it tries to tackle two difficult tasks simultaneously: consistent single-view novel view synthesis (NVS) and long-range extrapolation.
Consistent single-view novel view synthesis has been studied for a long time. 
Many methods~\cite{riegler2021stablevs, mildenhall2020nerf} propose utilizing multi-view data to infer the correspondences between frames, but they generally do not scale down to single- or few-view settings. 
Recently, there have also been attempts at single-view novel view synthesis. 
These methods mostly rely on learning a prior~\cite{yu2021pixelnerf, cai2022pix2nerf} or utilizing geometry information~\cite{xu2022sinnerf, wiles2020synsin, rockwell2021pixelsynth}. 
However, they do not generalize to long-range camera movement, as the content of the original image is quickly lost when taking large camera movements.
Current methods of long-range extrapolation~\cite{liu2020infnat, li2022infnat_zero, ren2022lotr, koh2021pathdreamer, rombach2021geofree} employ per-frame generation protocols, where the frames are generated in an auto-regressive feed-forward manner. 
The common downside of these methods is the lack of consistency between subsequent frames due to the per-frame refinement. 
A few recent methods~\cite{devries2021gsn, bautista2022gaudi} attempt to generate a whole scene directly using implicit representations.
However, this setting is computationally expensive, causing them to fail to achieve photo-realism even on low-resolution synthetic data.

Very recently, efforts have been made to perform scene extrapolation using pre-trained large-scale text-to-image diffusion models~\cite{ramesh2022dalle2, rombach2022latentdiffusion}.
This line of methods relies on prompt engineering and produces results with jittering between frames due to the lack of consistency enforcement.
However, image-conditioned diffusion models are naturally suitable for the task of scene extrapolation, as the guided denoising process can be interpreted as a search in the latent space.
Compared to feed-forward GAN-based methods~\cite{liu2020infnat, li2022infnat_zero, koh2021pathdreamer}, this allows the model to preserve high-level semantic meaning and low-frequency features while adding in high-frequency details and in-painting the missing parts.
We hence utilize these strengths of diffusion models for scene exploration and further improve its 3D consistency.

In this paper, we propose \textit{DiffDreamer}, a fully unsupervised method capable of consistent scene extrapolation given only a single image as input, and only internet photo collections as training data. Inspired by recent success in diffusion-based image refinement~\cite{lugmayr2022repaint, saharia2022palette}, we formulate consistent scene extrapolation as learning a conditional diffusion model from images only. We train the conditional diffusion model to generate the frames in an iterative refinement manner, showing that this allows convergence towards a harmonic set of frames with high fidelity. The consistency achieved by our conditional diffusion model potentially enables one to fuse the outputs as a 3D model, e.g., a NeRF~\cite{mildenhall2020nerf} with high consistency score~\cite{watson20223dim}. \edit{A key advantage of diffusion models is the flexibility of modifying their sampling behaviors at inference. By stochastic conditioning at inference, we can condition the generation on multiple past and future frames and form a bidirectional pipeline, despite having only single images during training.}

Experiments demonstrate that our framework allows one to synthesize a long-range fly-through sequence into an RGB image. We believe our framework not only serves as a starting point for consistent scene extrapolation and diffusion-model-based novel view synthesis but also scene extrapolation on more complex large-scale scenarios, such as autonomous driving scenes.

Our contributions include
\begin{itemize}
[leftmargin=*]
\setlength\itemsep{-.1em}
\item We introduce DiffDreamer, the first single-view scene extrapolation framework based on diffusion models for large-scale scenes.
\item We propose an anchored sampling strategy and a lookahead mechanism for long-range scene extrapolation. Combined with diffusion models, we significantly alleviate the well-known domain drifting issue~\cite{liu2020infnat, li2022infnat_zero} of scene extrapolation.
\item We demonstrate a fully automated scene-level novel view synthesis pipeline using conditional diffusion models.
%
\end{itemize}

%% file: sec/2_related.tex
\input{fig/nerfdreamer/overview.tex}
\section{Related works}
\label{sec:related}
\paragraph{Novel view synthesis from multi-view images}
Research related to Novel View Synthesis~(NVS) has a long history. Traditional multi-view NVS relies on inferring underlying geometry and interpolating the input images~\cite{penner2017soft3drecon, chaurasia2013depthsynth, debevec1998efficientibr, debevec1996modeling, fitzgibbon2005ibrusingibp, kopf2013ibr, chen1993viewinterp, gortler1996lumigraph, levoy1996light, seitz2006comparison, zitnick2004hqvideoviewinterp}. Recent successful attempts utilize deep learning methods to construct scene representations from multi-view data. These scene representations include but are not limited to: depth images~\cite{aliev2020npbg, meshry2019nrw, riegler2020fvs, yoon2020nvsds, tulsiani2018layerd3d, shih20203dphoto}, multi-plane images~\cite{tucker2020singlenvsmpi, zhou2018stereo}, voxels~\cite{liu2020nsvf, sitzmann2019deepvoxels}, and implicit functions~\cite{sitzmann2019srns, park2019deepsdf, peng2020convonet}. Among these representations, major progress has been made on radiance field approaches. Neural Radiance Fields~(NeRFs)~\cite{mildenhall2020nerf} have demonstrated encouraging progress for view synthesis by encoding color and transmittance in a multilayer perceptron, hence encoding a scene as an implicit representation. Using volumetric rendering, NeRF can perform photo-realistic novel view synthesis from only multi-view captured images and their poses. The outstanding performance of NeRF attracts tremendous efforts to improve its performance~\cite{barron2021mipnerf, barron2022mipnerf360, verbin2021refnerf, chen2022augnerf}, accelerate its training~\cite{mueller2022instant, yu_and_fridovichkeil2021plenoxels, sun2022dvgo, obukhov2022ttnf}, speed up rendering~\cite{autoint,sitzmann2021lfns}, extend or generalize it towards other downstream tasks~\cite{martinbrualla2020nerfw, mildenhall2021rawnerf, chan2020pigan, chan2021eg3d, schwarz2020graf, niemeyer2020giraffe, hao2021gancraft, zhu2022niceslam}, etc. Unlike these multi-view methods, we assume a single input image at the inference stage, where no geometry or interpolation can be inferred easily. Even with a single input image, our method can produce novel views faithfully.

\paragraph{Novel view synthesis from a single image}
There has been a vein of research on single-shot NVS. Some of them rely on geometry information or annotations~\cite{niklaus20193dkenburn, shin20193d, tulsiani2017factored3d}. However, geometry information and annotations are usually expensive to obtain for in-the-wild images. Other methods~\cite{chen2019mono, yu2021pixelnerf, cai2022pix2nerf, tulsiani2018layerd3d, sitzmann2019srns, jang2021codenerf, trevithick2021grf} relax this constraint by learning a prior filling in the missing information. These methods typically either only work well on simple objects~(e.g., ShapeNet~\cite{chang2015shapenet}) or restrict camera motions to small regions around the reference view. In contrast, we aim to relax such constraints and target long-range view extrapolation.

\paragraph{Scene extrapolation}
Long-range view extrapolation requires going beyond observations. With recent progress in generative modeling, several view extrapolation methods have emerged~\cite{brock2018large, chen2017photographic, ledig2017photo, isola2017pix2pix, karras2020analyzing, park2019spade, shaham2019singan, wang2018high, zhang2019self, rombach2021geofree}.  Earlier methods such as SynSin~\cite{wiles2020synsin} perform inpainting after reprojection, which struggles after a very limited range. A follow-up work, PixelSynth~\cite{rockwell2021pixelsynth}, works similarly to our DiffDreamer; it performs large-step image outpainting and accumulates a 3D point cloud for intermediate refinement and rendering. However, PixelSynth does not generalize to larger camera movements and requires a refinement module on top of the point cloud to enhance and inpaint, causing severely inconsistent intermediate view synthesis results.

\paragraph{Long-term path synthesis}
State-of-the-art methods such as InfNat~\cite{liu2020infnat}, InfNat-0~\cite{li2022infnat_zero}, PathDreamer~\cite{koh2021pathdreamer} and LOTR~\cite{ren2022lotr} deploy iterative training protocols and achieve perpetual view extrapolation for extremely long camera trajectories. However, these methods work in an auto-regressive per-frame generation framework. As a consequence, severe inconsistency can be observed from their rendered frames. Solving such inconsistency potentially requires generating an entire 3D world model, which is extremely computationally expensive, as shown by previous works~\cite{devries2021gsn, bautista2022gaudi}, whilst feed-forward per-frame generation methods~\cite{rombach2021geofree, liu2020infnat, li2022infnat_zero} suffer from content drifting and both local and global inconsistency. Therefore, we attempt to benefit from a diffusion-based iterative refinement method to generate consistent content.

\paragraph{Diffusion models}
The recent development of diffusion models~\cite{ho2020ddpm, sohl2015dulunt} has pushed AI-driven content creation to another level. These methods learn to transform any data distribution into a prior distribution, then sample new data by first sampling a random latent vector from the prior distribution, followed by ancestral sampling with the reverse Markov chain, parameterized by deep neural networks. The powerful diffusion/denoising mechanism enables various traditional image-based tasks, including super-resolution~\cite{saharia2022sr3, ho2022cdm}, inpainting~\cite{lugmayr2022repaint, saharia2022palette}, and editing~\cite{meng2021sdedit}. Their well-defined steady training protocols enhance the diffusion models' performance for large-scale training~\cite{rombach2022latentdiffusion, ramesh2022dalle2}. Very recently, success has been made to lift the strength of diffusion models to the 3D domain~\cite{poole2022dreamfusion, watson20223dim}, further demonstrating the potential of 3D-based diffusion models.

We formulate our task similarly to InfNat~\cite{liu2020infnat} and InfNat-0~\cite{li2022infnat_zero}, but use a conditional diffusion model instead of GANs and take focus on achieving consistency.

%% file: fig/nerfdreamer/overview.tex
\begin{figure*}
\begin{center}
\centering
\includegraphics[width=0.99\linewidth]{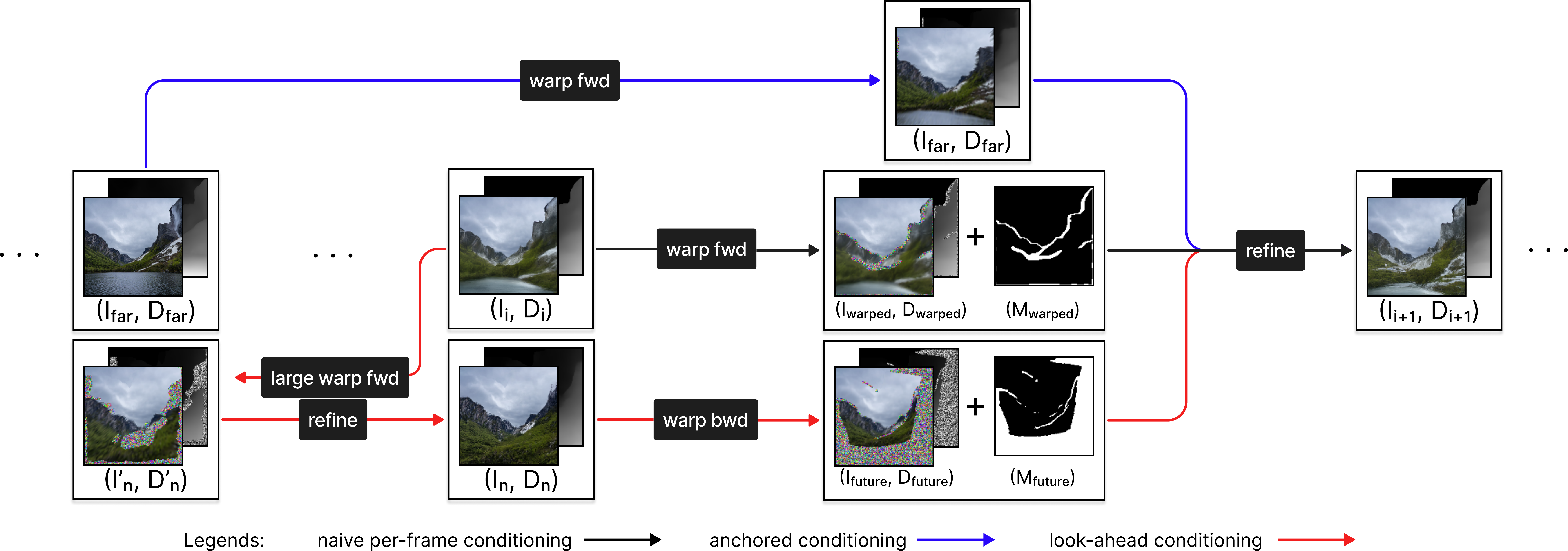}
\end{center}
\vspace{-15pt}
\caption{
\textbf{Overview of our pipeline}. We train an image-conditional diffusion model to perform image-to-image refinement and inpainting given a corrupted image and its missing region mask. At inference, we perform stochastic conditioning on three conditionings: naive forward warping from the previous frame~(black arrow), anchored conditioning by warping a further frame~(blue arrow), and lookahead conditioning by warping a virtual future frame~(red arrow). We repeat this render-refine-repeat pipeline to get sequences extrapolating a given image.
}
\label{fig:overview}
\end{figure*}

%% file: sec/3_method.tex
\section{DiffDreamer}
Given a single input image, the aim of DiffDreamer is to generate a consistent and harmonic 3D camera trajectory that represents flying into the given image. DiffDreamer addresses this task by training a conditional diffusion model to perform image inpainting and refinement concurrently. The overview of our pipeline is illustrated in Fig.~\ref{fig:overview}.

We synthesize frames of a fly-through video with three steps: \textit{render, refine and repeat}. In detail, given an RGB image $I_\mathrm{i}$ with its monocular predicted disparity $D_\mathrm{i}$ located at camera pose $c_\mathrm{i}$, we can unproject the colored pixels into 3D space and render the projected view at the next camera pose $c_\mathrm{i+1}$ by a 3D renderer~\cite{ravi2020pytorch3d} $\pi$: $(I_\mathrm{i+1}', D_\mathrm{i+1}')=\pi(I_\mathrm{i}, D_\mathrm{i}, c_\mathrm{i}, c_\mathrm{i+1})$. With a refinement network $F_\theta$, the warped RGBD image $(I_\mathrm{i+1}', D_\mathrm{i+1}')$ can be inpainted and refined to get a fine next frame $(I_\mathrm{i+1}$, $I_\mathrm{i+1}) = F_\theta(I_\mathrm{i+1}', D_\mathrm{i+1}')$, shown by the black arrow flow in Fig.~\ref{fig:overview}. We then treat $(I_\mathrm{i+1}$, $D_\mathrm{i+1})$ as the starting view of the next step and perform the warping and refinement stages repeatedly, yielding a set of frames extrapolating the scene.
\subsection{Training}
\label{sec:training}
Prior works~\cite{liu2020infnat, li2022infnat_zero} model the training process exactly as the render-refine-repeat pipeline since the process is fully differentiable. However, this naive approach is not generalizable to diffusion models for two reasons: 1) the training process of diffusion models is split into different noise levels, and 2) sampling from a diffusion model requires up to thousands of denoising steps. This means we need to store thousands of intermediate steps and gradients to perform back-propagation, which is computationally infeasible.

The main function of the ``repeat'' step during training is to feed the network with its own outputs to ameliorate
distribution drifting~\cite{liu2020infnat}. Therefore, it becomes critical to replace this step, especially when diffusion models are known to be sensitive to input distribution. Firstly, we create training pairs similar to \cite{li2022infnat_zero} by projecting a ground truth RGBD image $(I_\mathrm{gt}, D_\mathrm{gt})$ at initial camera pose $c_\mathrm{0}$ to a pseudo previous camera pose $c_\mathrm{pseudo}$: $(I_\mathrm{pseudo}, D_\mathrm{pseudo})=\pi(I_\mathrm{gt}, D_\mathrm{gt}, c_\mathrm{0}, c_\mathrm{pseudo})$, then project back to $c_\mathrm{0}$ to create a corresponding corrupted RGBD: $(I_\mathrm{corrupted}, D_\mathrm{corrupted})=\pi(I_\mathrm{pseudo}, D_\mathrm{pseudo}, c_\mathrm{pseudo}, c_\mathrm{0}))$. We thus obtain a pair of ground truth RGBD images $(I_\mathrm{gt}, D_\mathrm{gt})$ and its corrupted version $(I_\mathrm{corrupted}, D_\mathrm{corrupted})$, which involves missing parts and warping artifacts simulating a forward motion.

Having these paired data enables training an image-conditioned diffusion model  $p(\bm{y} | \bm{x}, \bm{m})$, where $\bm{x}=(I_\mathrm{corrupted}, D_\mathrm{corrupted})$, a corrupted version of ground truth image $\bm{y}=(I_\mathrm{gt}, D_\mathrm{gt})$ while $\bm{m}$ denotes the missing region mask from warping. We train the model with the following objective~\cite{ho2020ddpm, saharia2022palette}:
\begin{equation}
    L(\theta) = 
    \mathbb{E}_{(\bm{x}, \bm{m}, \bm{y})} \mathbb{E}_{\bm{\epsilon} } \mathbb{E}_{\gamma}\, \left[ \left\|  f_\theta(\bm{x}, \bm{m},\, {\widetilde{\bm{y}}}, \,\gamma) - \bm{\epsilon} \right\|^{2}_2 \right] ,
\label{eq:train_loss}
\end{equation}
where $\widetilde{\bm{y}} = \sqrt{\gamma} \,\bm{y} + \sqrt{1\!-\!\gamma}\, \bm{\epsilon}~, \bm{\epsilon} \sim \mathcal{N}(\bm{0},\bm{I})~$, and $\gamma$ indicates the noise level. Note that since the diffusion model needs to concurrently learn inpainting and refinement, we additionally condition the neural network on the missing region mask to provide stronger guidance, following prior works~\cite{liu2020infnat, li2022infnat_zero}.
Similar to previous work \cite{li2022infnat_zero}, we assume that the sky region lies infinitely far away and does not change. Therefore, we inject noise only to the ground region to obtain $\widetilde{\bm{y}}$.

\subsection{Inference}
Diffusion models trained as described above perform well for a single forward step but do not generalize to long-term due to severe domain drifting after only a few iterations. This causes the extrapolated results to gradually drift away after only a handful of steps~(see Sec.~\ref{sec:ablation_studies}). We propose two strategies at the inference stage to counter this issue and preserve both local and global consistency.

\input{fig/results/qualitative_results.tex}

\subsubsection{Anchored conditioning}
We introduce anchored conditioning, which conditions the diffusion model on long-range camera movements in order to enhance consistency over larger distances.
As shown in prior work~\cite{watson20223dim}, it is feasible to naively approximate true auto-regressive sampling via stochastic conditioning for conditional diffusion models. While moving forward from camera pose $c_\mathrm{i}$ to $c_\mathrm{i+1}$, instead of strictly conditioning on the warped previous image $(I_\mathrm{warped}, D_\mathrm{warped})=\pi(I_\mathrm{i}, D_\mathrm{i}, c_\mathrm{i}, c_\mathrm{i+1}))$ and mask during the inference denoising stage, we additionally select a frame $(I_\mathrm{far}, D_\mathrm{far})$ at a previous camera pose $c_\mathrm{far}$ further away from the current camera position. In practice, we empirically select the current frame $(I_\mathrm{i}, D_\mathrm{i})$ every 5 steps. We then perform stochastic conditioning on the warped previous frame $(I_\mathrm{warped}, D_\mathrm{warped})$ and an ``anchored'' frame by warping $(I_\mathrm{far}, D_\mathrm{far})$ to the desired camera pose: $(I_\mathrm{anchored}, D_\mathrm{anchored})=\pi(I_\mathrm{far}, D_\mathrm{far}, c_\mathrm{far}, c_\mathrm{i+1})$. We perform this conditioning without specifying the missing region mask, as anchored conditioning requires longer-range warping, which may introduce more regions as missing, thus undermining the goal of long-term consistency. Conditioning on the warped previous frame naively encourages frame-to-frame consistency, while conditioning on a far-away frame offers long-term consistency. Stochastic conditioning on a largely warped image is also helpful with respect to domain drifting, as it is easier to simulate the same artifacts and blurriness during training by simply warping ground truth images equally further away. Thanks to the diffusion models' steady training protocol, refining and inpainting largely warped images with massive missing areas can be learned jointly during training.

\subsubsection{Virtual lookahead conditioning}
Prior works~\cite{liu2020infnat, li2022infnat_zero} deploy per-frame generation; therefore, they suffer from severe inconsistency. A straightforward approach to solve this is to generate a scene representation directly, but this is extremely challenging and expensive to perform. While anchored conditioning solves parts of the global consistency issue, warping an image distorts and stretches the texture, and blurs out fine details.

We notice that compared with flying into an image and refining the artifacts and blurriness, it is significantly easier to zoom out of an image and outpaint the missing regions without suffering from domain drift. This is due to the available regions preserving high-frequency details, which confer a strong signal for filling in missing regions.

Therefore, adding in a ``lookahead'' mechanism in diffusion models is helpful for both achieving long-term consistency and preventing domain drifting, as we can benefit from conditioning the generation on a future image, whose fine details preserve after warping to the current pose. While flying deep into an image, we observe that the generated content shares little overlap with the input image. Utilizing this fact, we can create a virtual view lying ahead for a sequence of views. With stochastic conditioning, this is as simple as additionally conditioning the generation on $(I_\mathrm{future}, D_\mathrm{future})$, acquired by warping a shared virtual view $(I_\mathrm{n}, D_\mathrm{n})$ lying ahead at a shared virtual future camera pose $c_\mathrm{n}$ warped to each camera pose $c_\mathrm{i}$: $(I_\mathrm{future}, D_\mathrm{future})=\pi(I_\mathrm{n}, D_\mathrm{n}, c_\mathrm{n}, c_\mathrm{i})$. In practice, we empirically generate $c_\mathrm{n}$ by taking a forward motion 10 times larger than a single step. We update the shared $(I_\mathrm{n}, D_\mathrm{n})$ every 10 steps and condition the future 10 frames on it. The shared virtual view can be flexibly generated by refining an available view to a future camera pose, a randomly generated view, or even another real image. To preserve consistency, we find it is suitable to warp the current frame $(I_\mathrm{i}, D_\mathrm{i})$ to a future camera pose $c_\mathrm{n}$ significantly beyond a single forward motion so that there is enough ambiguity, then refine to get virtual lookahead conditioning $(I_\mathrm{future}, D_\mathrm{future})$.

With the proposed anchored and virtual lookahead conditioning, we can formulate each denoising step going from camera pose $c_\mathrm{i}$ to $c_\mathrm{i+1}$ as:
\begin{equation}
\begin{split}
\bm{y}_{t-1} \leftarrow & \frac{1}{\sqrt{\alpha_t}} \left( \bm{y}_t - \frac{1-\alpha_t}{ \sqrt{1 - \gamma_t}} f_{\theta}(\bm{x}_\mathrm{i}, \bm{m}_\mathrm{i}, \bm{y}_{t}, \gamma_t) \right) \\
& + \sqrt{1 - \alpha_t}\bm{\epsilon}_t~
\end{split}
\end{equation}
at the inference stage, where $\bm{x}_\mathrm{i}$ is a weighted selection of (0.5, 0.25, 0.25) among $(I_\mathrm{warped}, D_\mathrm{warped})$, $(I_\mathrm{anchored}, D_\mathrm{anchored})$ and $(I_\mathrm{future}, D_\mathrm{future})$, with $\bm{m}_\mathrm{i}$ being the missing region mask. We apply classifier-free guidance~\cite{ho2021cfg} during inference as it encourages the denoising process to take more signals from the conditioning.

\subsection{Training details}
We carefully design the training protocols according to our inference strategies. Firstly, instead of assuming a fixed step size like previous works~\cite{liu2020infnat, li2022infnat_zero}, we randomly choose step size from a range $(-\mathrm{s}, \mathrm{s})$ while training the model to generalize to long-range conditioning, where we empirically choose $\mathrm{s}=20$. Note that we also train the model to fly out of the image for the purpose of lookahead conditioning. We find injecting random Gaussian noise into the missing regions rather than preserving the stretched details or masking out the missing regions to be very helpful, as it serves as an additional latent space that encourages diversity and effectively reduces the domain drifting between forward motions and circular motions we used for creating pseudo training pairs. To add support for classifier-free guidance~\cite{ho2022classifier} at inference, we zero out all conditioning inputs with 10\% probability during training.

%% file: fig/results/qualitative_results.tex
\begin{figure*}
\begin{center}
\centering
\includegraphics[width=0.99\linewidth, clip=true, trim = 0cm 0cm 0cm 0cm]{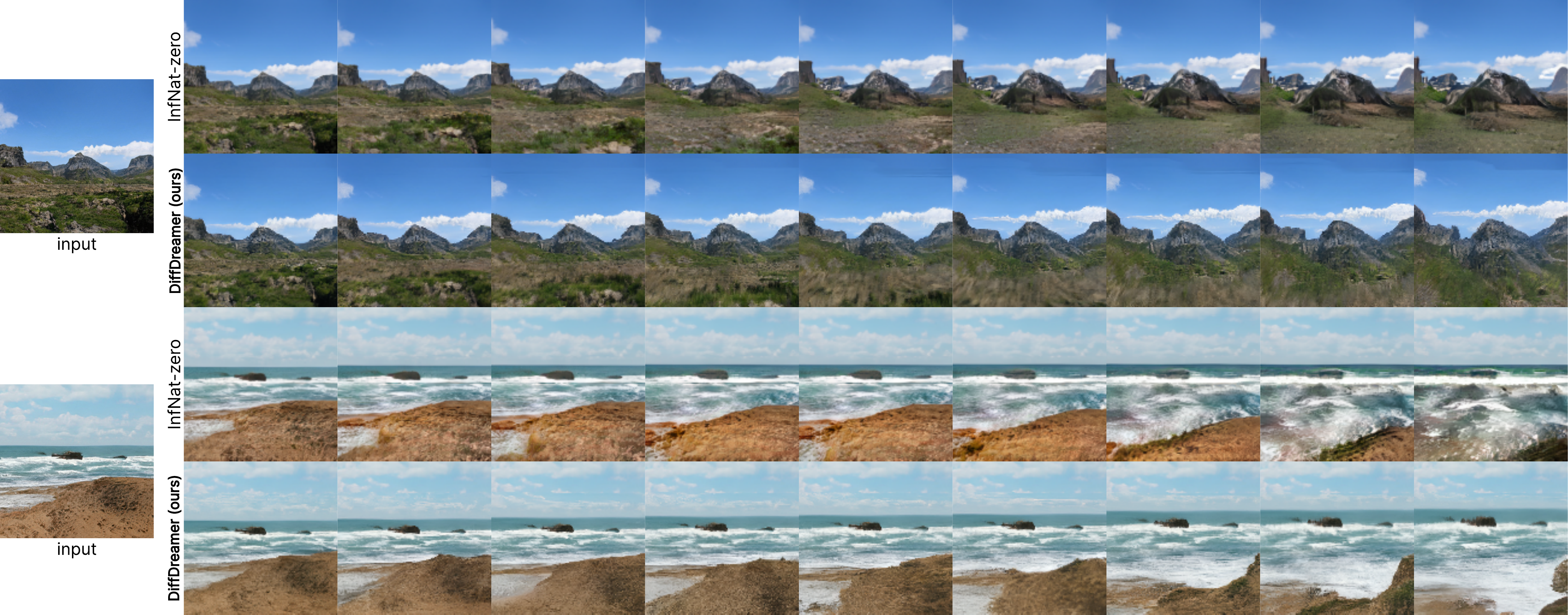}
\end{center}
\vspace{-15pt}
\caption{
\textbf{Qualitative comparisons of InfNat-0~\cite{li2022infnat_zero} and our DiffDreamer generation}, for which we ask the models to fly toward a target region and compare the outputs. Note that as InfNat-0~\cite{li2022infnat_zero} is not 3D consistent and may need more steps even with identical input disparities and camera speed, we manually inserted more refinement steps to our DiffDreamer to ensure it is a fair comparison. Even so, we do not observe significant drifting from our DiffDreamer, while InfNat-0~\cite{li2022infnat_zero} is incapable of preserving the input domain.
}
\label{fig:qualitative_results}
\end{figure*}

%% file: sec/4_results.tex
\input{fig/results/perpetual_view_generation.tex}
\input{fig/results/consistency_comparison.tex}
\input{fig/results/detail_zoom_in.tex}

\section{Experiments}
\label{sec:experiments}
\subsection{Evaluation}
\paragraph{Datasets}
We report results on the LHQ~\cite{skorokhodov2021lhq} dataset, a collection of 90K nature landscape photos.
Following prior work~\cite{li2022infnat_zero}, we use the full data for training and 100 images provided by \cite{li2022infnat_zero}'s authors, generated from a pre-trained StyleGAN2~\cite{karras2020stylegan2ada},  as the test set.
\edit{Following \cite{liu2020infnat}, we also supply quantitative results on the ACID~\cite{liu2020infnat} dataset, with evaluation on 50 input images from its test set.}

\paragraph{Evaluation metrics}
Evaluation of scene extrapolation frameworks is non-trivial as there is no single evaluation metric that covers every aspect of the generation quality.
We follow the evaluation protocol of prior works~\cite{liu2020infnat, li2022infnat_zero} on the rendered sequences.
For that, we report Inception Score (IS)~\cite{salimans2016is}, Frechet Inception Distance~(FID)~\cite{heusel2017fid} and Kernel Inception Distance~(KID)~\cite{binkowski2018kid} with scaling factor $\times$10 computed using the torch-fidelity package~\cite{obukhov2020torchfidelity}. 
We evaluate the models in two settings: a shorter range of 20 refinement steps\edit{, a middle range of 50 steps,} and a longer range of 100 refinement steps. 
We additionally report 3D consistency scoring~\cite{watson20223dim}, a recent metric for evaluating 3D consistency. We compute this metric by generating a sequence of frames from an input, training a neural field~\cite{mildenhall2020nerf} with a fraction of the generated frames, and calculating PSNR~(Peak Signal to Noise Ratio), SSIM~(Structural Similarity Index Measure)~\cite{zhou2004ssim} and Perceptual Similarity~(LPIPS)~\cite{zhang2018lpips} against the held-out generated frames. We evaluate 3D consistency scoring over 10 sequences of 30 frames from 10 randomly selected input images.
As our prior works~\cite{liu2020infnat, li2022infnat_zero} also output disparity maps, we use a disparity supervised DVGOv2~\cite{sun2022dvgo} as the underlying neural field model. Following \cite{watson20223dim}, we zero out the viewing direction conditioning to avoid overfitting to view-dependencies.
To further assess local consistency, we compare the number of points from COLMAP~\cite{schoenberger2016sfm, schoenberger2016mvs} reconstruction on the same 10 rendered video sequences, using the default automatic reconstruction without specifying the camera poses.

\paragraph{Qualitative results}
Qualitative comparisons are shown in Fig.~\ref{fig:qualitative_results}, where we ask the rendering models to fly toward a target. We compare the intermediate frames rendered by the model. Interestingly, we find that even with identical input depth and step size, it takes significantly more refinement steps for InfNat-0~\cite{li2022infnat_zero} to reach a target due to a tendency to render the far plane to be further than the mesh projection. In contrast, our DiffDreamer maintains high 3D consistency, and because it takes fewer steps to reach a target, it exhibits less drift. However, to ensure a fair comparison with~\cite{li2022infnat_zero} and to evaluate our model's drift under more refinement steps, we manually insert additional intermediate camera poses between each pair of nearby autocruise~\cite{liu2020infnat} poses so that the models conduct exactly the same number of refinement steps to reach the final frame.

We compare DiffDreamer's consistency against InfNat-0 in Fig.~\ref{fig:consistency_comparison}, along with a reference mesh projection created by warping the initial image into the final view according to the estimated disparity map. With identical disparity map and step size provided by InfNat-0~\cite{li2022infnat_zero}'s authors, our model shows significantly better alignment with the projected mesh. While InfNat-0~\cite{li2022infnat_zero} renders decent intermediate frames, it demonstrates only limited consistency with the mesh. Although it accurately models the expected movement of the foreground, the hill in the mid-distance remains static, whereas we expect it to fill the frame as the camera flies forward. This artifact may be due to a bias that encourages maintaining useful distant content while training scene extrapolation models. By enforcing the lookahead mechanism, which enforces future frames to be consistent with the mesh projection, our DiffDreamer does not suffer from this issue. We show an example of the detail-preserving ability of DiffDreamer in Fig~\ref{fig:detail_zoom_in}.

We additionally visualize examples of scene extrapolation of 50 steps in Fig.~\ref{fig:perpetual_view_generation}. Despite only seeing single images during training, the learned generative prior enables DiffDreamer to perform long-range extrapolation.

\paragraph{Quantitative results}
\input{tab/lhq.tex}
\input{tab/acid.tex}
We show the quantitative evaluation on LHQ~\cite{skorokhodov2021lhq} in Tab.~\ref{tab:lhq_results} \edit{and ACID~\cite{liu2020infnat} in Tab.~\ref{tab:acid_results}}. We observe that our 20-step generation outperforms prior works on all metrics by a relatively large margin. \edit{Our 50-step generation also has a significant advantage over prior works except for InfNat~\cite{liu2020infnat} and InfNat-0~\cite{li2022infnat_zero}, which are on par with our model.}
\input{tab/3d_consistency_score.tex}
\input{tab/colmap_reconstruction.tex}
Our DiffDreamer's 100-step generation is not as good as \cite{liu2020infnat} and \cite{li2022infnat_zero} on FID~\cite{heusel2017fid} and KID~\cite{binkowski2018kid} while achieving higher IS~\cite{salimans2016is}. However, we achieve significantly better 3D consistency metrics, as shown in Tab.~\ref{tab:3d_consistency_score} and Tab.~\ref{tab:colmap_recon}. Our model performs scene extrapolation based on the presented content from the input image very well, but we do not enforce it to generate diverse content. Therefore, the model may output blander frames after it goes significantly beyond the input. We also notice that our better 3D consistency makes autocruise~\cite{liu2020infnat} fail and hit the ground/hills more often, constituting a large portion of the failed scenes. This is due to our model preserving geometry cues—it does not refine regions to be further away from their actual positions. \edit{Note that though InfNat~\cite{liu2020infnat} and InfNat-0~\cite{li2022infnat_zero} can synthesize sequences with hundreds of frames, they resemble complete trade-offs with consistency. We argue that without accurate geometry preservation, scene extrapolation models~\cite{liu2020infnat, li2022infnat_zero} will converge towards random latent space walk using pretrained GANs.}

\subsection{Ablation studies}
\label{sec:ablation_studies}
\input{fig/results/ablation_study.tex}
We perform a thorough ablation study to verify our design choices, shown in Fig.~\ref{fig:ablation_study}. The setups are: 1) Naive auto-regressive: we first set up a baseline by performing the simplest per-step generation. The naive method fails after only a few refinement steps. This is due to the input domain drifting as observed in prior works~\cite{liu2020infnat, li2022infnat_zero}. 2) Without anchored conditioning: next, we disable anchored conditioning: we observe more severe 3D inconsistency while moving forward, as the conditioning signal is purely from the past frame. 3) Without lookahead conditioning: we proceed with removing the lookahead mechanism. We observe significant domain drifting and artifacts as the model no longer takes advantage of the easier flying-out task. 
\edit{We supply the corresponding quantitative ablations in Appendix Sec.~\ref{sec:quantitative_ablation}.}

%% file: fig/results/perpetual_view_generation.tex
\begin{figure*}
\begin{center}
\centering
\includegraphics[width=0.99\linewidth]{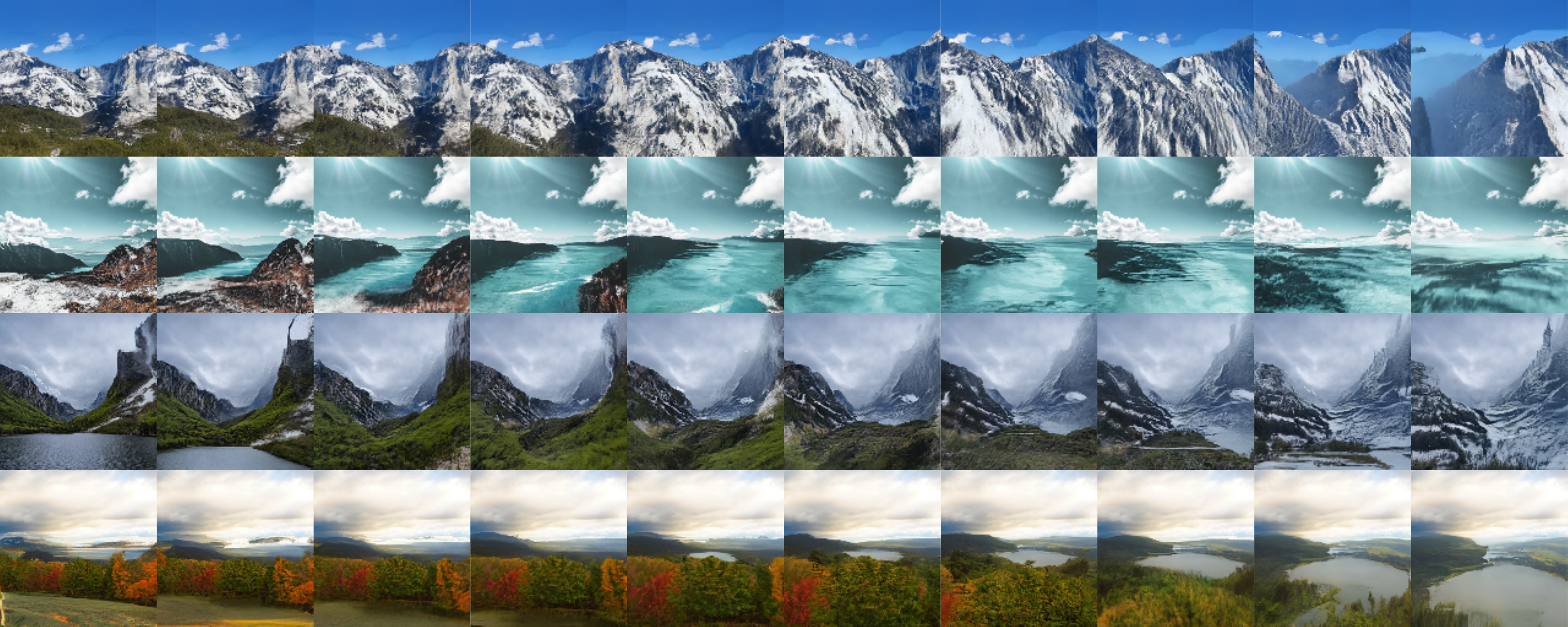}
\end{center}
\vspace{-15pt}
\caption{
\textbf{Long-range view extrapolation} of over 50 steps forward.
}
\label{fig:perpetual_view_generation}
\end{figure*}

%% file: fig/results/consistency_comparison.tex
\begin{figure*}
\begin{center}
\centering
\includegraphics[width=0.99\linewidth, clip=true, trim = 6.4cm 1.7cm 0cm 6cm]{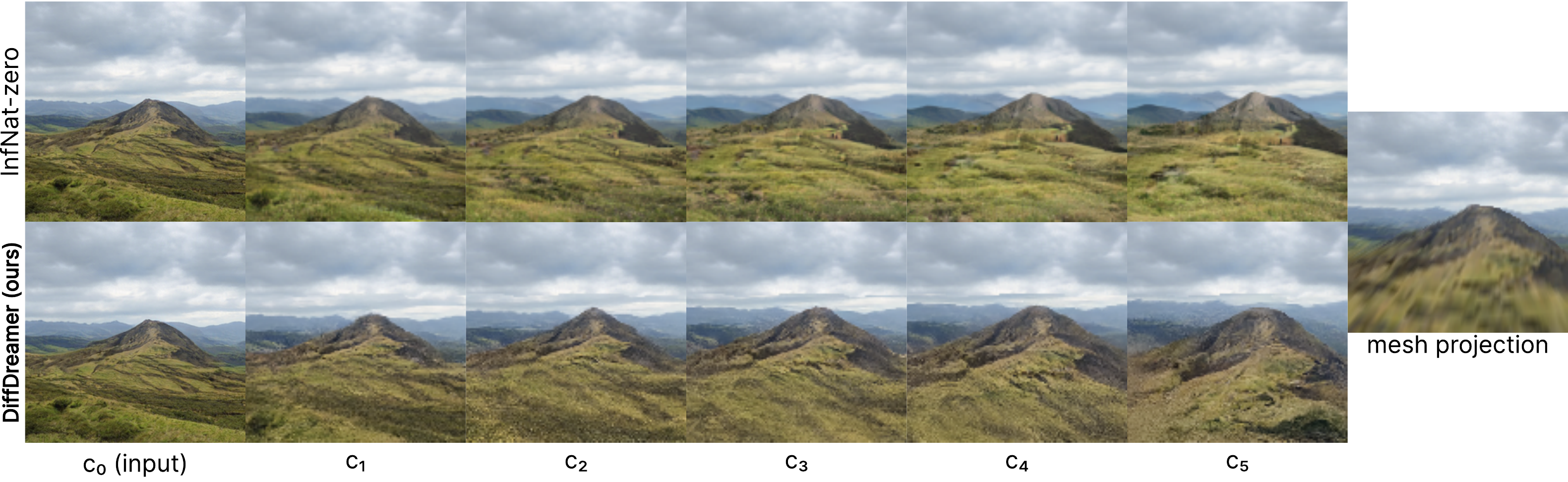}
\end{center}
\vspace{-15pt}
\caption{
\textbf{Comparison of 3D consistency} achieved by our DiffDreamer and InfNat-0~\cite{li2022infnat_zero}, where we ask the camera to fly towards the top of the hill and show the intermediate renderings at camera positions $c_\mathrm{0}$ to $c_\mathrm{5}$.
}
\label{fig:consistency_comparison}
\end{figure*}

%% file: fig/results/detail_zoom_in.tex
\begin{figure*}
\begin{center}
\centering
\includegraphics[width=0.99\linewidth]{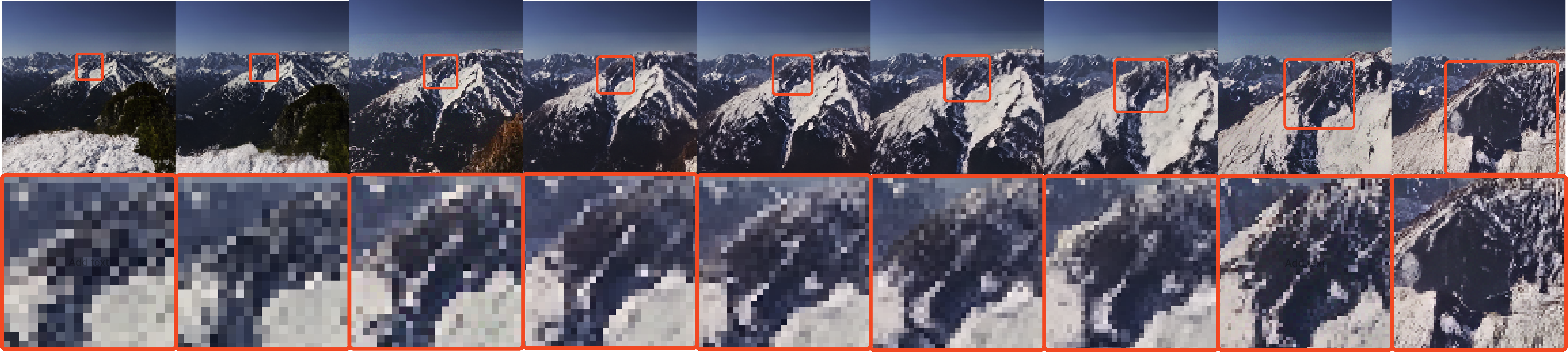}
\end{center}
\vspace{-15pt}
\caption{
\textbf{Detail preservation} of over 30 steps. DiffDreamer can preserve details upon long-range extrapolation.
}
\label{fig:detail_zoom_in}
\end{figure*}

%% file: tab/lhq.tex
\begin{table}[t!]

\begin{center}
\small
\resizebox{0.99\linewidth}{!}{
\begin{tabular}{@{}lccc|ccc|ccc@{}}
\toprule
& \multicolumn{3}{c}{first 20 steps} & \multicolumn{3}{c}{first 50 steps} & \multicolumn{3}{c}{full 100 steps}\\
Method & FID$\downarrow$ & KID$\downarrow$ & IS$\uparrow$ & FID$\downarrow$ & KID$\downarrow$ & IS$\uparrow$ & FID$\downarrow$ & KID$\downarrow$ & IS$\uparrow$ \\ 
\midrule
InfNat-0~\cite{li2022infnat_zero}
& 39.45 & 0.12 & 2.80& \textbf{36.53} & \textbf{0.11} & 2.79 & \textbf{26.24} & \textbf{0.12} & 2.72 
\\
\midrule
DiffDreamer
& \textbf{34.49} & \textbf{0.08} & \textbf{2.82} & 38.86 & 0.12 & \textbf{2.90} & 51.0 & 0.28 & \textbf{2.99} 
\\
\bottomrule
\end{tabular}
}
\vspace{-5pt}
\caption{\textbf{Quantitative results on LHQ~\cite{skorokhodov2021lhq}} for 20 steps, 50 steps, and 100 steps generation. To our knowledge, InfNat-0~\cite{li2022infnat_zero} is the only prior work capable of long-range view synthesis without supervision from sequential data or accurate ground truth depth.}
\label{tab:lhq_results} 
\end{center} 
\vspace{-5pt}
\end{table}

%% file: tab/acid.tex
\begin{table}[t!]
\centering
\resizebox{\linewidth}{!}{ 
\begin{tabular}{@{}lccc|ccc|ccc@{}}
\toprule
& \multicolumn{3}{c}{first 20 steps} & \multicolumn{3}{c}{ first 50 steps} & \multicolumn{3}{c}{full 100 steps} \\
Method & FID $\downarrow$ & KID $\downarrow$ & IS $\uparrow$ & FID $\downarrow$ & KID $\downarrow$ & IS $\uparrow$ & FID $\downarrow$ & KID $\downarrow$ & IS $\uparrow$\\
\midrule
SynSin~\cite{wiles2020synsin} & 79.58 & 0.63 & 1.90 & 96.37 & 0.78 & 1.71 & 108.95 & 1.06 & 1.75 \\
PixelSynth~\cite{rockwell2021pixelsynth} & 89.63 & 1.10 & 1.23 & 97.14 & 1.32 & 1.42 & 107.61 & 1.20 & 1.63 \\
3D Photos~\cite{shih20203dphoto} & 99.79 & 0.80 & 1.65 & 123.60 & 0.79 & 1.12 & 111.39 & 0.87 & 1.58 \\
InfNat~\cite{liu2020infnat} & 59.93 & 0.22 & 2.36 & \textbf{57.47} & \textbf{0.26} & 2.28 & \textbf{48.27} & \textbf{0.27} & 2.28\\
\midrule
DiffDreamer & \textbf{52.81} & \textbf{0.12} & \textbf{2.69} & 61.04 & \textbf{0.26} & \textbf{2.86} & 70.11 & 0.41 & \textbf{2.82}\\
\bottomrule
\end{tabular}
} 
\vspace{-5pt}
\caption{
\textbf{Quantitative results on ACID~\cite{liu2020infnat}} for 20 steps, 50 steps, and 100 steps generation. Note that all prior works require \textit{posed} multi-view sequences for training, while our DiffDreamer is trained from single image collections.
} 
\label{tab:acid_results}
\end{table}%

%% file: tab/3d_consistency_score.tex
\begin{table}

\begin{center}
\small
\resizebox{0.85\linewidth}{!}{
\begin{tabular}{c|ccc}
\toprule
Method & PSNR$\uparrow$ & SSIM$\uparrow$ & LPIPS$\downarrow$
\\
\midrule
InfNat~\cite{liu2020infnat} & 19.94$\rpm$1.63 & 0.55$\rpm$0.07 & 0.18$\rpm$0.04 \\
InfNat-0~\cite{li2022infnat_zero} & 18.92$\rpm$1.42 & 0.41$\rpm$0.08 & 0.20$\rpm$0.02
\\
\midrule
DiffDreamer & \textbf{23.56}$\rpm$3.30 & \textbf{0.68}$\rpm$0.04 & \textbf{0.12}$\rpm$0.02
\\
\bottomrule
\end{tabular}
}
\vspace{-5pt}
\caption{\textbf{3D consistency scoring~\cite{watson20223dim}}, where we train disparity supervised DVGOv2~\cite{sun2022dvgo} using 10 sequences generated by the models and report the mean$\rpm$std novel view synthesis metrics.}
\label{tab:3d_consistency_score}  
\end{center}
\vspace{-20pt}
\end{table}

%% file: tab/colmap_reconstruction.tex
\begin{table}

\begin{center}
\small
\resizebox{0.655\linewidth}{!}{
\begin{tabular}{cc}
\toprule
Method & Avg. points reconstructed\\ 
\midrule
InfNat~\cite{liu2020infnat} & 1476$\rpm$477 \\
InfNat-0~\cite{li2022infnat_zero} & 612$\rpm$104
\\
\midrule
DiffDreamer & \textbf{3124}$\rpm$622
\\
\bottomrule
\end{tabular}
}
\vspace{-5pt}
\caption{\textbf{Number of reconstructed points via COLMAP~\cite{schoenberger2016sfm, schoenberger2016mvs}}, where we run COLMAP on 10 generated sequences, count the number of reconstructed points and report mean$\rpm$std.}
\label{tab:colmap_recon}  
\end{center}
\vspace{-20pt}
\end{table}

%% file: fig/results/ablation_study.tex
\begin{figure}
\vspace{-0.2cm}
\small
\begin{center}
{\small
\hspace*{-0.25cm}
\includegraphics[width=0.99\columnwidth]{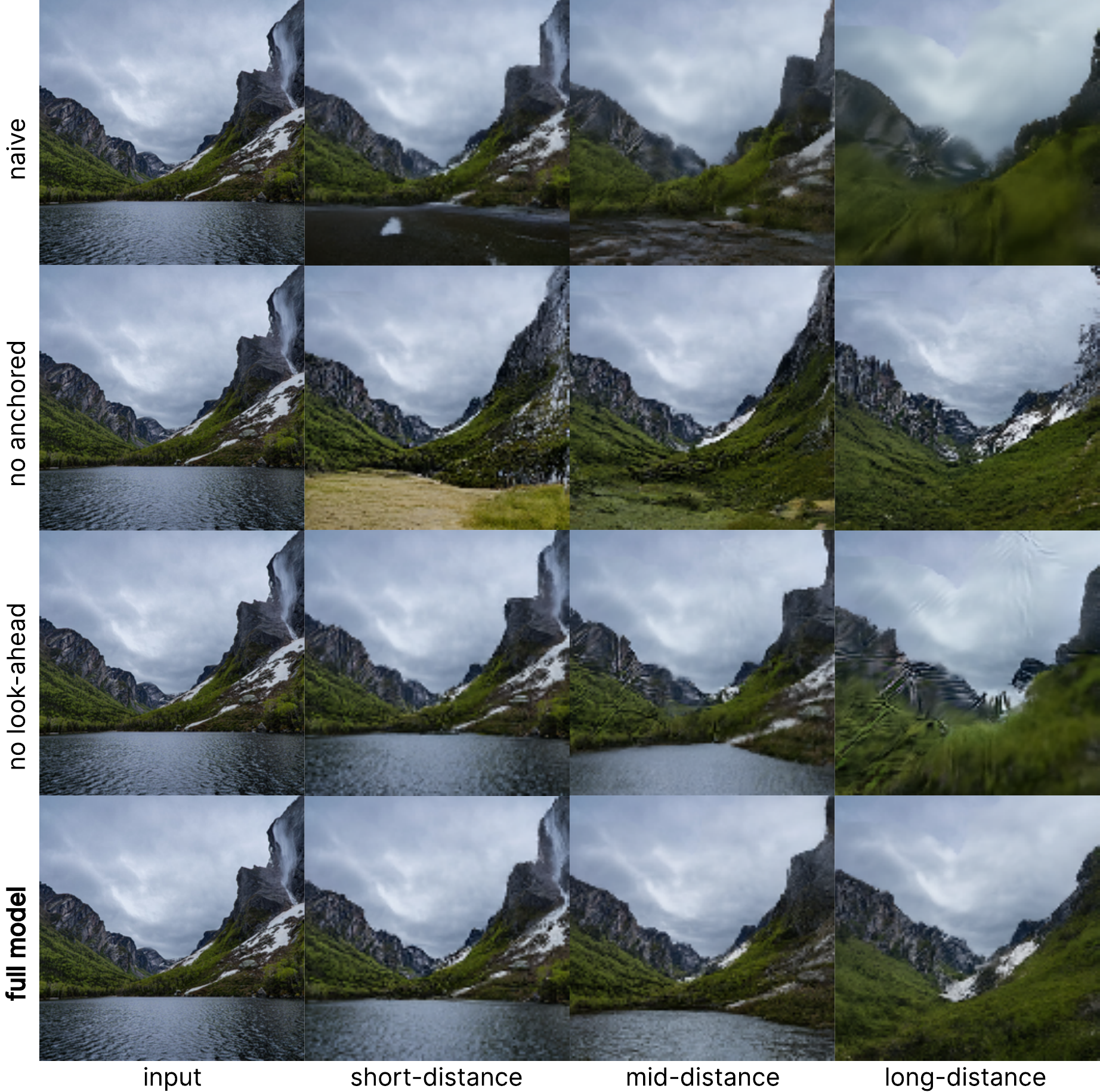}
}
\end{center}
\vspace{-15pt}
\caption{\textbf{Ablation study}, where we disabled our key building blocks. We observe clear artifacts~(naive, no lookahead) or inconsistency~(no anchored) in all ablations.
}
\vspace{-10pt}
\label{fig:ablation_study}
\end{figure}

%% file: sec/5_conclusions.tex
\section{Discussion}
In this paper, we introduced DiffDreamer, a novel unsupervised pipeline based on conditional diffusion models for scene extrapolation. Diffdreamer can conduct scene extrapolation capable of ``flying'' into the image while training only from internet-collected single images. The key idea of DiffDreamer is to utilize a conditional diffusion model to simultaneously inpaint and refine a corrupted image obtained by warping a previous image.
This is accomplished by training a conditional diffusion model that is capable of performing image-to-image translation under various corruption augmentations and utilizing stochastic conditioning to refine a corrupted image given multiple conditioning.
Our model demonstrated comparable generation quality with GAN-based methods while maintaining significantly better consistency than prior works.

\paragraph{Limitation and future work}
DiffDreamer cannot synthesize novel views in real time due to the heavy inference of diffusion models. However, speeding up diffusion models' inference has been a very active area, and we expect advances in this area to speed up our approach directly.
In addition, we do not enforce the diversity of content while going significantly beyond the input image, causing degradation for a true perpetual generation. We believe adding CLIP~\cite{radford2021clip} conditioning to be a very exciting extension.

\paragraph{Conclusion}
Diffusion models are emerging as state-of-the-art generative 2D methods. DiffDreamer is the first approach to apply them to 3D scene extrapolation, demonstrating a high amount of view consistency that is crucial for many downstream tasks.

\paragraph{Acknowledgement}
We thank Zhengqi Li for providing all the technical details and data samples related to InfNat-0~\cite{li2022infnat_zero}. Gordon Wetzstein was supported by Samsung, Stanford HAI, and a PECASE from the ARO.

%% file: sec/X_supplementary.tex
\appendix
\balance

\twocolumn[
\centering
\Large
\textbf{DiffDreamer: Towards Consistent Unsupervised Single-view Scene Extrapolation with Conditional Diffusion Models} \\
\vspace{0.5em}Supplementary Material \\
\vspace{1.0em}
] 
\appendix

\section{Additional qualitative results}\label{sec:additional_qualitative_results}
Figures ~\ref{fig:additional_qualitative_results_2}, \ref{fig:additional_qualitative_results_3}, \ref{fig:additional_qualitative_results_4}, and \ref{fig:additional_qualitative_results_1} show additional scene extrapolation results from our model with 50 steps of forward motion. 
The task of scene extrapolation has a multi-modal nature: given a single input image, there could be infinite ways of generation. Therefore, we show multiple rendering trajectories of over 50 steps for each input image and supporting videos with framerates upsampled using~\cite{reda2022film}~(note that the videos are rendered at 128$\times$128 and may appear blurry under higher resolution). To encourage diversity and prevent hitting mountains/the ground while generating longer sequences, we can additionally condition the diffusion model on randomly selected patterns from the input image while generating the pseudo future frames with a weight of 0.2. We select these patterns by simply performing free-form brush stroke masking, using the algorithm provided in \cite{yu2018free, yu2018generative} and refer to this diversity-focused version as ``DiffDreamer-diverse'', encouraging diversity over long-range at a price of trading-off consistency. We supply results for this diversity-focused setting and show frames from a generated 500-step sequence in Fig.~\ref{fig:500_frame}.
\input{fig/results/500_frame.tex}
\input{fig/results/fail_case.tex}

\section{Additional quantitative results}\label{sec:quantitative_ablation}
We supply quantitative ablation comparisons including DiffDreamer-diverse on LHQ~\cite{skorokhodov2021lhq} in Tab.~\ref{tab:ablation}, and additional quantitative results of DiffDreamer-diverse on ACID~\cite{liu2020infnat} in Tab.~\ref{tab:additional_quantitative}.
\input{tab/ablation}
\input{tab/additional_quantitative}

\section{Flying-out}
Even though we do not design our model specifically for flying-out setting, DiffDreamer has a significant advantage over naïve autoregression. Since we are working on outdoor scenes, dramatic depth discontinuities will appear~\cite{fridman2023scenescape}. This is especially obvious when the flying-out motion is not just a straight translation. Our bi-directional method is a good counter to this issue since future frame guidance and simultaneous refinement can alleviate the artifacts. We show example flying-out sequences in Fig.~\ref{fig:flying_out} and include accompanying videos with 100 steps.

\section{Technical details}
We use the U-Net backbone from \cite{dhariwal2021guided_diffusion} and train all models for 1M iterations with a mini-batch size of 128. We trained our model for roughly a week and 3 days respectively for LHQ~\cite{skorokhodov2021lhq} and ACID~\cite{liu2020infnat}, on 2 NVIDIA RTX 8000 GPUs. We compare against the released pretrained InfNat and InfNat-zero models, which were trained for 8 days on 10 GPUs, and 6 days on 8 GPUs respectively. We build our model on top of Palette~\cite{saharia2022palette} and use the Adam optimizer with a learning rate of 1e-4 and a 10k linear learning rate warm-up schedule. We also employ 0.9999 EMA for our model. During both training and inference, we use a linear noise schedule of (1e-6, 0.01) with 2000 time steps.
Following prior works~\cite{liu2020infnat, li2022infnat_zero}, we extract monocular-predicted disparity maps with MiDaS~\cite{ranftl2022midas}, and sky region masks using DeepLab~\cite{liang2017deeplab}.
We adopt the autocruise algorithm from \cite{liu2020infnat} to sample the camera path for both training and inference. The autocruise algorithm uses the disparity map to estimate the skyline and horizon,  then generate a camera trajectory that avoids hitting the ground or hills. We follow \cite{li2022infnat_zero} during inference and use a camera speed of 0.1875. We train and evaluate our model on image resolution of 128$\times$128 to be consistent with prior work~\cite{li2022infnat_zero}.
\section{Autocruise specifics}
We use the autocruise algorithm from \cite{liu2020infnat} to generate camera trajectories for both training and evaluation. As we only have raw images as training data, whose intrinsics are unknown and cannot be easily inferred, we follow \cite{li2022infnat_zero} and randomly sample the field of view~(FoV) between 45$^{\circ}$ and 70$^{\circ}$, and fix to 55$^{\circ}$ during testing. Autocruise algorithm deploys a mechanism to predict the next camera pose by encouraging the next view to have a $\tau_\mathrm{sky}$ fraction of sky regions~(determined by thresholding disparity less than 0.08) and a fraction of $\tau_\mathrm{near}$ fraction of nearby regions~(determined by thresholding disparity larger than 0.4). We follow \cite{li2022infnat_zero} to uniformly sample $\tau_\mathrm{near}$ from [0.2, 0.4] and $\tau_\mathrm{sky}$ from [0.25, 0.45] during training, and fix them to be 0.25 and 0.1 respectively during inference. In contrast to \cite{liu2020infnat, li2022infnat_zero}, which only moves a small fraction $\tau_\mathrm{lerp}=0.05$ of the way to the target directions at each frame to ensure smooth camera pose changing, we only use $\tau_\mathrm{lerp}=0.05$ during inference of our next frame and increase $\tau_\mathrm{lerp}=0.3$ for generating the pseudo future frame. We uniformly sample $\tau_\mathrm{lerp}$ from [0.0, 0.3] during training. We direct readers to \cite{liu2020infnat, li2022infnat_zero} for further specifics of the autocruise algorithm.
\section{Mesh renderer specifics}
We use a PyTorch implementation~\cite{shi2020ptmeshrenderer} of a 3D mesh renderer~\cite{genova2018tfmeshrenderer}. Following \cite{liu2020infnat}, each pixel is projected into the 3D space using its disparity and is then treated as a vertex connected with its neighbors to form a triangle mesh. To obtain the missing region masks, we follow \cite{liu2020infnat} and threshold the gradient of the input disparity by 0.3 to make a mask, which refers to the regions with sharp disparity change. We project the mask to target the camera pose to get the final missing region mask.
\section{Dataset pre-processing}
Both of the LHQ~\cite{skorokhodov2021lhq} dataset 
and the ACID~\cite{liu2020infnat} dataset contains many samples unsuitable for training scene extrapolation models. This includes images focusing on the foreground and images of the ground, with camera poses pointed downward. Following \cite{li2022infnat_zero}, we filter out images whose minimum MiDaS~\cite{ranftl2022midas} predicted disparity value is larger than 200.
\section{Failure cases}
There are two main causes for failures. First, we do not enforce diversity of outputs. During training, the model always sees real images. This means during our pseudo pairs generation, the corrupted version of the ground truth image will still be diverse, even if it is under a lower frequency due to warping artifacts. However, while we are going significantly beyond the input image's content, any future frame will solely rely on the model's outputs, which may not exhibit enough diverse content for moving forward. We show an example of this case in Fig.~\ref{fig:fail_case}. We believe it is exciting to extend DiffDreamer to support vector conditioning, e.g., CLIP embedding conditioning, to enforce output diversity.

Second, as our model has significantly better geometry alignment than \cite{liu2020infnat}, the autocruise algorithm fails more often, causing the camera trajectory to hit mountains or the ground, despite our best efforts in tuning its parameters. We show an example of this failure case in Fig.~\ref{fig:fail_case}.

\input{fig/results/flying_out}
\input{fig/results/additional_qualitative_results_2.tex}
\input{fig/results/additional_qualitative_results_3.tex}
\input{fig/results/additional_qualitative_results_4.tex}
\input{fig/results/additional_qualitative_results_1.tex}

%% file: fig/results/500_frame.tex
\begin{figure*}
\begin{center}
\centering
\includegraphics[width=0.99\linewidth]{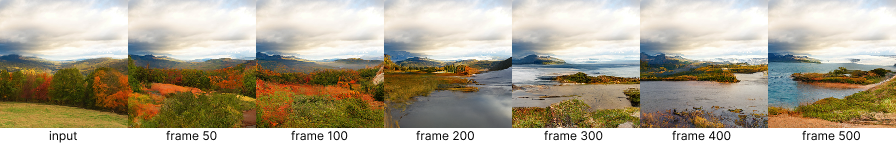}
\end{center}
\vspace{-15pt}
\caption{
\textbf{Perpetual view synthesis} of a sequence over 500 steps.
}
\label{fig:500_frame}
\end{figure*}

%% file: fig/results/fail_case.tex
\begin{figure*}
\begin{center}
\centering
\includegraphics[width=0.4\columnwidth]{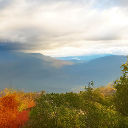}
\includegraphics[width=0.4\columnwidth]{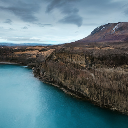}
\end{center}
\vspace{-15pt}
\caption{
\textbf{Failure cases} when the model's output is not diverse enough to support future frames~(left) or the autocruise algorithm gets too close to the mountains/ground~(right).
}
\label{fig:fail_case}
\end{figure*}

%% file: tab/ablation.tex
\begin{table}[H]
\centering
\resizebox{\linewidth}{!}{ 
\begin{tabular}{@{}lccc|ccc|ccc|c@{}}
\toprule
& \multicolumn{3}{c}{20 steps} & \multicolumn{3}{c}{50 steps} & \multicolumn{3}{c}{100 steps} & COLMAP\\
Method & FID $\downarrow$ & KID $\downarrow$ & IS $\uparrow$ & FID $\downarrow$ & KID $\downarrow$ & IS $\uparrow$ & FID $\downarrow$ & KID $\downarrow$ & IS $\uparrow$ &\\
\midrule
InfNat-0 & 39.45 & 0.12 & 2.80 & 36.53 & 0.11 & 2.79 & 26.24 & \textbf{0.12} & 2.72 & 612 \\
\midrule
Auto-regressive & 70.53 & 0.53 & 1.99 & 77.81 & 0.63 & 1.91 & 90.69 & 0.81 & 2.14 & 2030 \\
No anchored & 38.41 & 0.17 & 2.70 & 46.40 & 0.24 & 2.63 & 58.67 & 0.40 & 2.79 & 1543 \\
No lookahead & 68.30 & 0.46 & 1.76 & 75.18 & 0.74 & 1.85 & 92.85 & 0.81 & 2.06 & 2457 \\
\midrule
DiffDreamer & \textbf{34.49} & \textbf{0.08} & 2.82 & 38.86 & 0.12 & 2.90 & 51.0 & 0.28 & 2.99 & \textbf{3124} \\
DiffDreamer-diverse & 34.92 & 0.09 & \textbf{3.19} & \textbf{30.78} & \textbf{0.10} & \textbf{3.27} & \textbf{24.04} & \textbf{0.12} & \textbf{3.26} & 1403 \\
\bottomrule
\end{tabular}
} 
\vspace{-5pt}
\caption{
Quantitative ablation studies.
} 
\label{tab:ablation}
\end{table}%

%% file: tab/additional_quantitative.tex
\begin{table}[H]
\vspace{-3mm}
\centering
\resizebox{\linewidth}{!}{ 
\begin{tabular}{@{}lccc|ccc|ccc|c@{}}
\toprule
& \multicolumn{3}{c}{20 steps} & \multicolumn{3}{c}{50 steps} & \multicolumn{3}{c}{100 steps} & COLMAP \\
Method & FID $\downarrow$ & KID $\downarrow$ & IS $\uparrow$ & FID $\downarrow$ & KID $\downarrow$ & IS $\uparrow$ & FID $\downarrow$ & KID $\downarrow$ & IS $\uparrow$ & \\
\midrule
InfNat & 59.93 & 0.22 & 2.36 & 57.47 & 0.26 & 2.28 & 48.27 & 0.27 & 2.28 & 1476\\
\midrule
DiffDreamer & 52.81 & \textbf{0.12} & \textbf{2.69} & 61.04 & 0.26 & \textbf{2.86} & 70.11 & 0.41 & \textbf{2.82} & \textbf{3423} \\
DiffDreamer-diverse & \textbf{51.28} & 0.15 & 2.37 & \textbf{44.44} & \textbf{0.19} & 2.40 & \textbf{42.97} & \textbf{0.21} & 2.43 & 1883 \\
\bottomrule
\end{tabular}
} 
\vspace{-5pt}
\caption{
Quantitative comparison of DiffDreamer-diverse's performance on ACID~\cite{liu2020infnat}.
} 
\label{tab:additional_quantitative}
\end{table}%

%% file: fig/results/flying_out.tex
\begin{figure*}
\begin{center}
\centering
\includegraphics[width=0.99\linewidth]{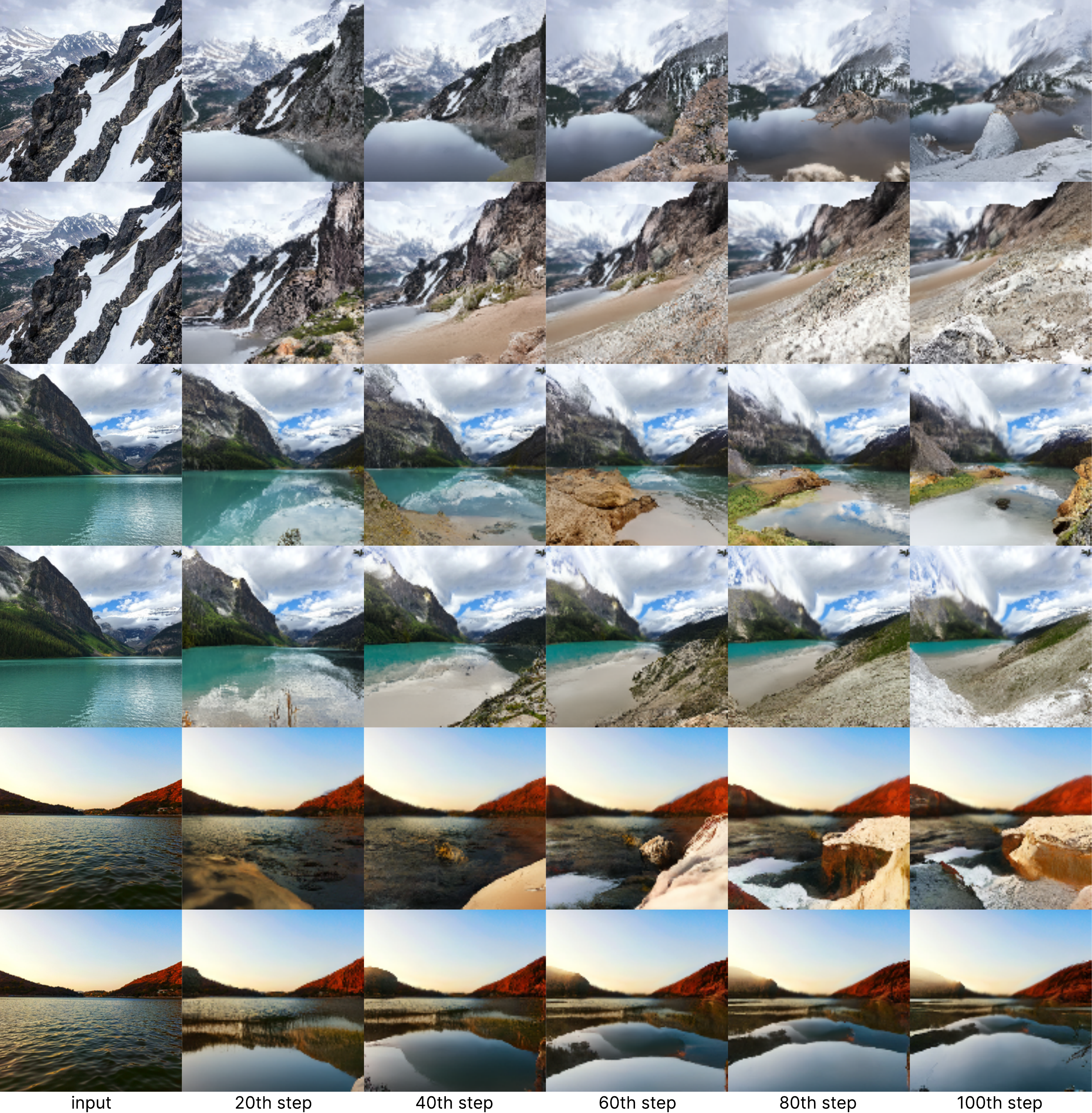}
\end{center}
\vspace{-15pt}
\caption{
\textbf{Flying-out} 100 steps of the input images.
}
\label{fig:flying_out}
\end{figure*}

%% file: fig/results/additional_qualitative_results_2.tex
\begin{figure*}
\begin{center}
\centering
\includegraphics[width=0.99\linewidth]{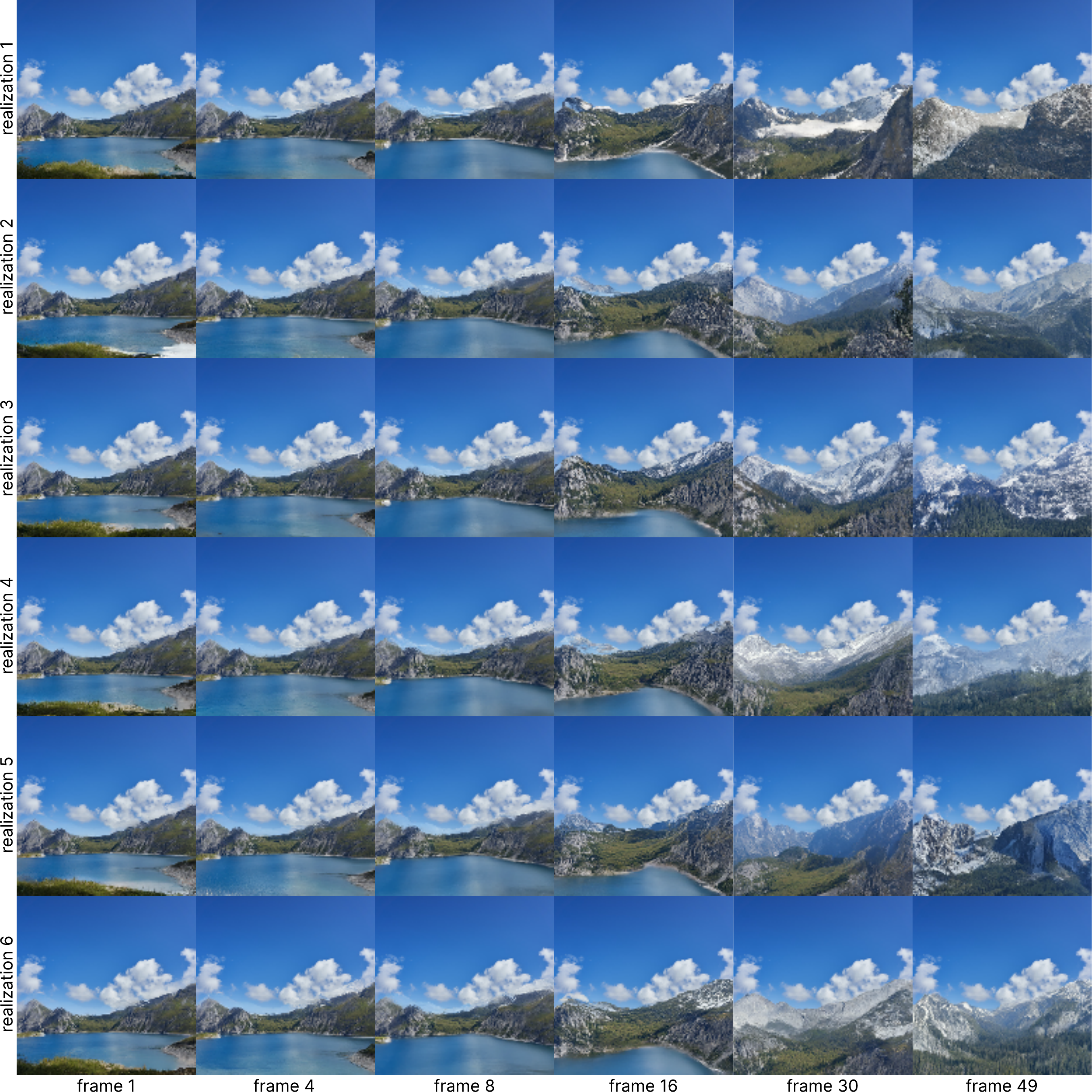}
\end{center}
\caption{
\textbf{Additional qualitative results:} Six distinct realizations, synthesized over 50 steps of forward motion.
}
\label{fig:additional_qualitative_results_2}
\end{figure*}

%% file: fig/results/additional_qualitative_results_3.tex
\begin{figure*}
\begin{center}
\centering
\includegraphics[width=0.99\linewidth]{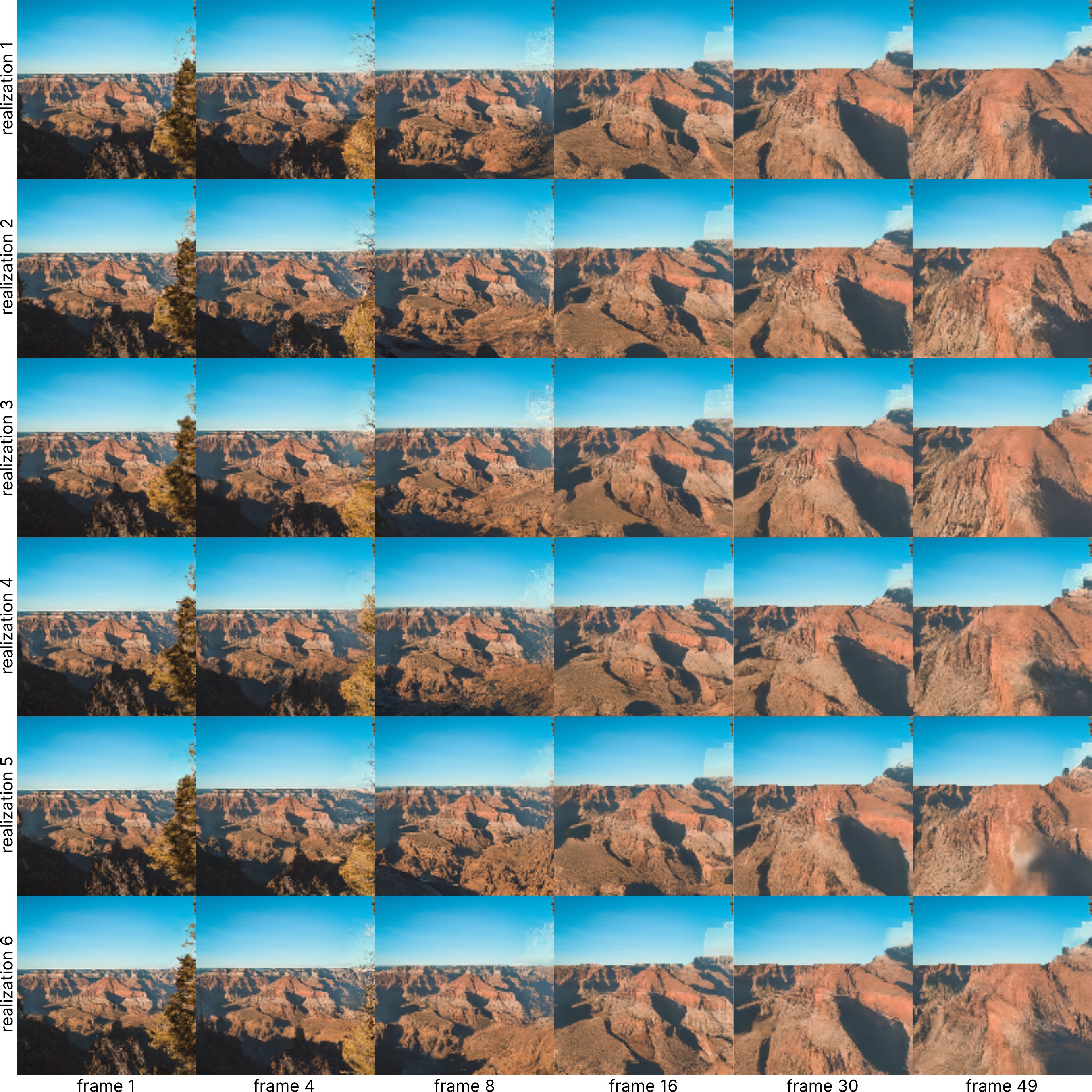}
\end{center}
\caption{
\textbf\textbf{Additional qualitative results:} Six distinct realizations, synthesized over 50 steps of forward motion.
}
\label{fig:additional_qualitative_results_3}
\end{figure*}

%% file: fig/results/additional_qualitative_results_4.tex
\begin{figure*}
\begin{center}
\centering
\includegraphics[width=0.99\linewidth]{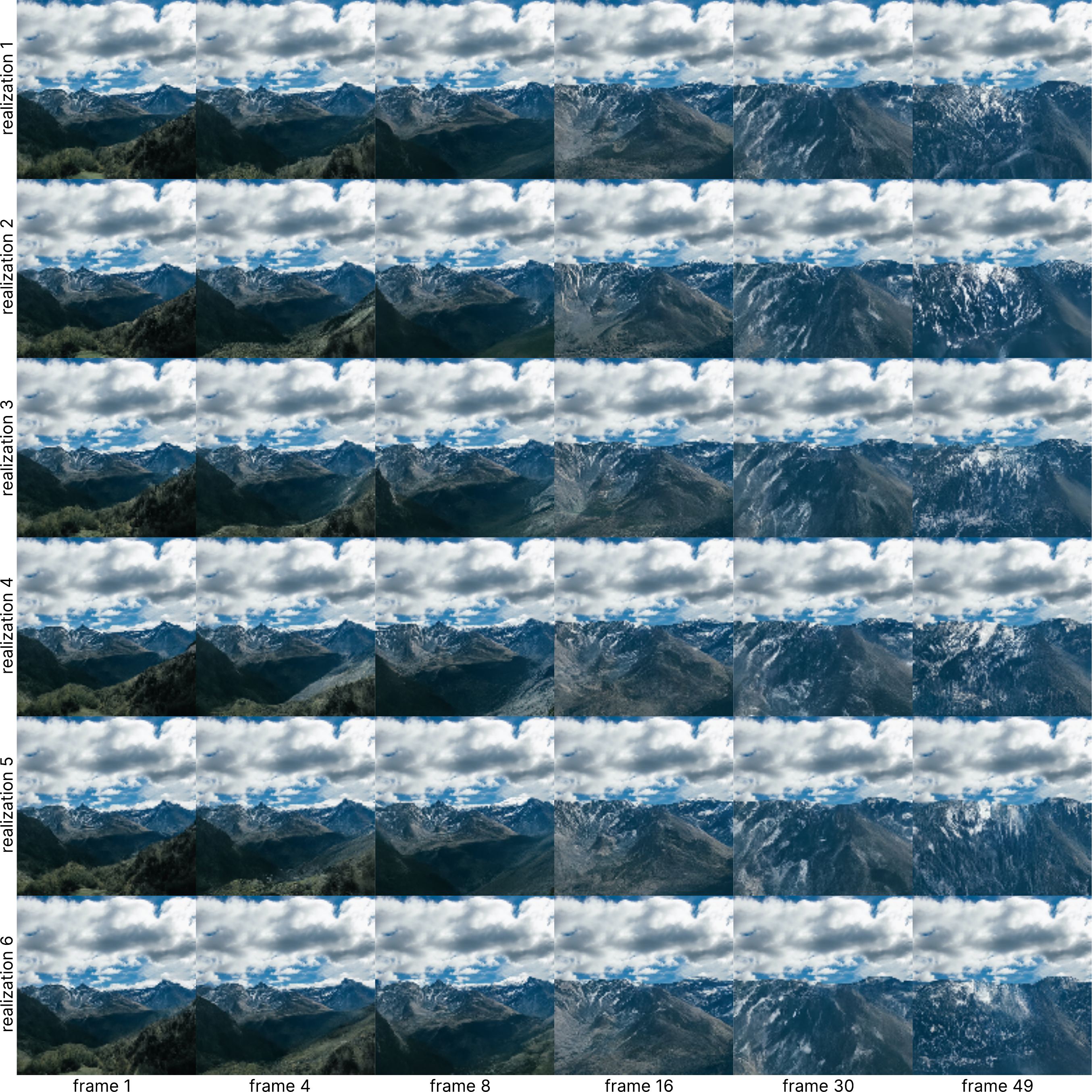}
\end{center}
\caption{
\textbf{Additional qualitative results:} Six distinct realizations, synthesized over 50 steps of forward motion. DiffDreamer is able to preserve consistency when there is no significant refinement needed.
}
\label{fig:additional_qualitative_results_4}
\end{figure*}

%% file: fig/results/additional_qualitative_results_1.tex
\begin{figure*}
\begin{center}
\centering
\includegraphics[width=0.99\linewidth]{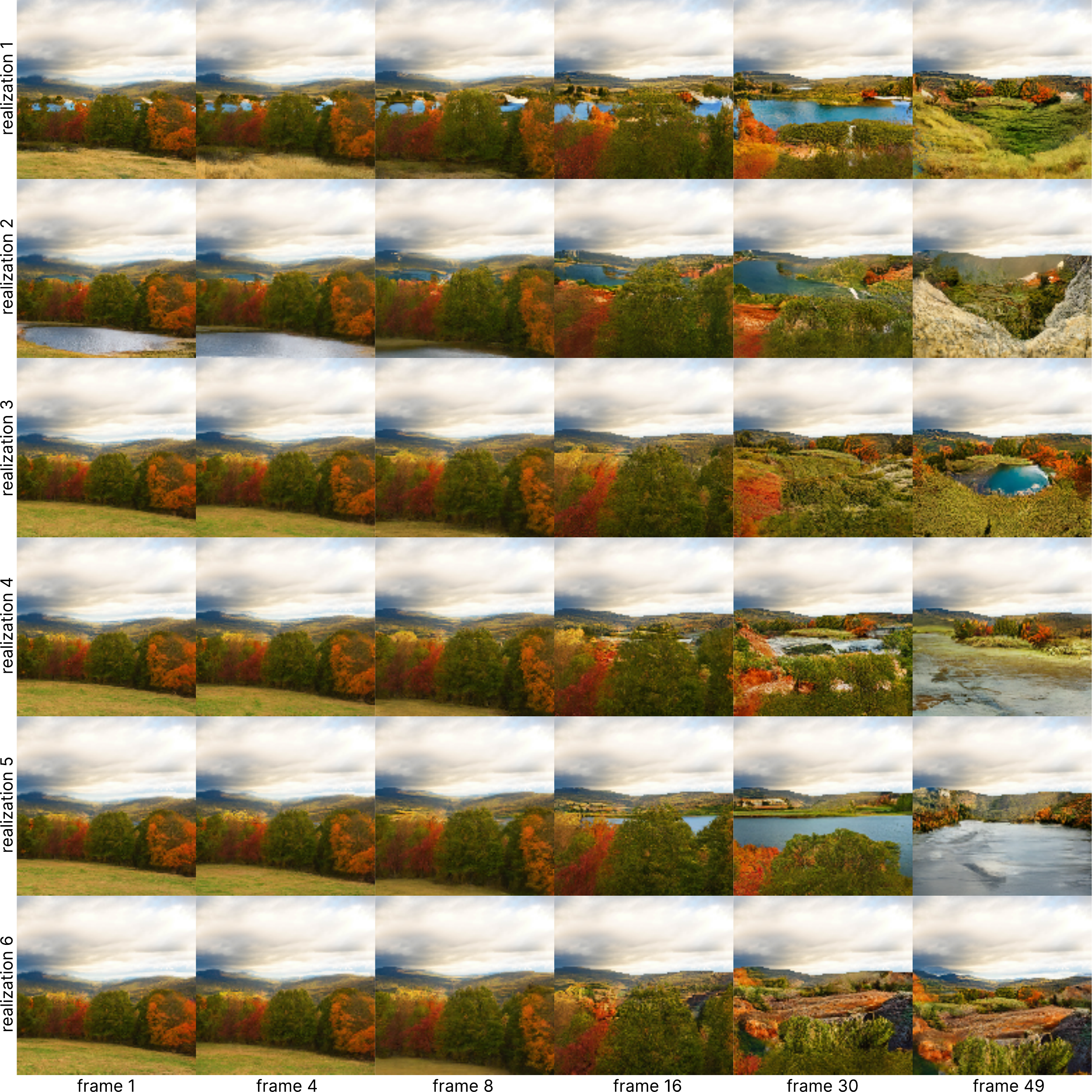}
\end{center}
\caption{
\textbf{Additional qualitative results:} Six distinct realizations, synthesized over 50 steps of forward motion, where we encourage output diversity by additionally conditioning on input patterns.
}
\label{fig:additional_qualitative_results_1}
\end{figure*}